\documentclass{article}

\usepackage{microtype}
\usepackage{graphicx}
\usepackage{subcaption}
\usepackage{booktabs} 

\usepackage{hyperref}

\usepackage[accepted]{icml2026}

\usepackage{amsmath}
\usepackage{amssymb}
\usepackage{mathtools}
\usepackage{amsthm}

\usepackage[capitalize,noabbrev]{cleveref}

\theoremstyle{plain}

\theoremstyle{definition}

\theoremstyle{remark}

\usepackage[textsize=tiny]{todonotes}

\usepackage{booktabs}
\usepackage{tabularx}
\usepackage{array}
\usepackage{makecell}
\usepackage{float}

\usepackage{svg}
\usepackage{amsfonts}
\usepackage{multirow}
\usepackage{amsmath}

\icmltitlerunning{Submission and Formatting Instructions for ICML 2026}

\begin{document}

\twocolumn[
  \icmltitle{MemBoost: A Memory-Boosted Framework for Cost-Aware LLM Inference}



  \icmlsetsymbol{equal}{*}

  \begin{icmlauthorlist}
    \icmlauthor{Joris Köster}{equal,yyy}
    \icmlauthor{Zixuan Liu}{equal,comp}
    \icmlauthor{Siavash Khajavi}{sch}
    \icmlauthor{Zizhan Zheng}{comp}
  \end{icmlauthorlist}

  \icmlaffiliation{yyy}{Department of Computer Science, Aalto University, Finland}
  \icmlaffiliation{comp}{Department of Computer Science, Tulane University, LA, USA}
  \icmlaffiliation{sch}{Department of Industrial Engineering and Management, Aalto University, Finland}

  \icmlcorrespondingauthor{Zixuan Liu}{zliu41@tulane.edu}

  \icmlkeywords{Machine Learning, ICML}

  \vskip 0.3in
]



\printAffiliationsAndNotice{\icmlEqualContribution}

\begin{abstract}
  Large Language Models (LLMs) deliver strong performance but incur high inference cost in real-world services, especially under workloads with repeated or near-duplicate queries across users and sessions. In this work, we propose \textbf{MemBoost}, a memory-boosted LLM serving framework that enables a lightweight model to reuse previously generated answers and retrieve relevant supporting information for cheap inference, while selectively escalating difficult or uncertain queries to a stronger model. Unlike standard retrieval-augmented generation, which primarily grounds a single response, MemBoost is designed for interactive settings by supporting answer reuse, continual memory growth, and cost-aware routing. Experiments across multiple models under simulated workloads show that MemBoost substantially reduces expensive large-model invocations and overall inference cost, while maintaining high answer quality comparable to the strong model baseline.
\end{abstract}

\section{Introduction}

Large Language Models (LLMs)~\cite{hurst2024gpt,team2023gemini} have demonstrated strong capabilities in natural language understanding~\cite{liu2025llava}, instruction following~\cite{chung2024scaling,ouyang2022training,liu2024enhancing}, coding~\cite{gao2023pal,ni2024l2ceval}, and decision-making~\cite{anil2023palm,liu2026targeting,liu2026ua,wei2022chain}. However, deploying frontier-scale models in real-world services remains expensive, as inference often requires multiple high-end GPUs, especially for long-form reasoning and explanation-heavy responses~\cite{kwon2023efficient,park2025survey}. These costs are amplified in production settings where many user queries are repeated or near duplicates, causing redundant computation. A widely adopted direction to mitigate these costs is to shift part of the workload to smaller models with retrieval-augmented generation (RAG)~\cite{lewis2020retrieval,guu2020retrieval,borgeaud2022improving}. RAG allows a model to externalize knowledge into a searchable index and condition generation on retrieved evidence, reducing the need to store everything in parameters~\cite{izacard2023atlas,fan2025minirag}. For example, Atlas~\cite{izacard2023atlas} demonstrates that the RAG model with far fewer parameters can be competitive with, and in some regimes outperform, much larger models on knowledge-intensive tasks. 

However, most existing RAG work primarily targets knowledge grounding, i.e., improving the answer to a single query by retrieving external documents~\cite{lewis2020retrieval,guu2020retrieval}. In contrast, interactive LLM services exhibit additional properties that are not addressed by standard RAG alone. First, many queries that are semantically equivalent are repeatedly requested across users and sessions. Recomputing these same answers wastes GPU time. This motivates semantic caching, which retrieves a previously generated response when a new query is semantically similar~\cite{gill2025meancache,yu2025smartcache,liu2025semantic}. However, making this reliable requires careful control. Second, deployed LLMs continuously generate new high-quality answers. When these answers are valuable, they should be written back into the retrieval memory so that a semantic cache can serve future similar queries cheaply. However, this write-back mechanism raises practical questions, such as what/when to store it. Finally, although small models augmented with retrieval can be competitive on knowledge-intensive tasks, their capabilities remain limited on more challenging requests. A practical system should therefore enable the small model to escalate to a stronger model when retrieval is insufficient, aligning with recent work on routing across multiple LLMs~\cite{ong2024routellm,zhang2025router,moslem2026dynamic} to balance quality and cost.

These observations suggest a missing system-level paradigm: a unified approach that enables a small model to cheaply reuse supporting knowledge or prior answers from semantic memory, selectively defer to a stronger but costly model when retrieval is insufficient, and continuously write back useful new information into the retrieval memory, while explicitly balancing answer quality and inference cost. In this paper, we introduce \textbf{MemBoost}, a memory-boosted architecture with three components: (1) an Associative Memory Engine (AME) that performs fast semantic retrieval and supports write-back of newly generated answers; (2) a high-capability Large-LLM Oracle that provides an accurate fallback when memory is insufficient; and (3) a lightweight Meta Controller, which is a small LLM that composes the final response by either reusing from memory or routing the query to the oracle, and support write back mechanism for future reuse, thereby balancing quality and cost. Together, MemBoost turns inference into a “retrieve-or-escalate” decision problem, substantially reducing expensive large-model calls while preserving answer quality. Experiments on the MMLU-Pro dataset~\cite{wang2024mmlupro} under a simulated workload with repeated and near-duplicate queries show that MemBoost substantially reduces calls to the costly large LLM while largely preserving the answer quality of the oracle model.

\section{MemBoost: Memory-Boosted LLM Serving Framework}

\subsection{Problem Setup}

Unlike standard single-turn question answering (QA) tasks studied in prior work, we consider a setting with continuous interaction between multiple users and an LLM service, where incoming queries may be exact duplicates or semantic near-duplicates over time. At each time step $t$, the system receives a user query $x_t$ and produces an answer $y_t$. We use a ground-truth quality signal $r(x_t,y_t)$ to measure whether the response is helpful, correct, and aligned with the user’s intent. Our objective is to maximize the average response quality over a horizon $T$, \(
\max \frac{1}{T}\sum_{t=1}^{T}r(x_t,y_t),
\) while minimizing the overall inference cost, as formalized in Section~\ref{sec:cost}.

\subsection{Overview of MemBoost}

\begin{figure}[!t]
    \centering
    \includegraphics[width=0.85\linewidth]{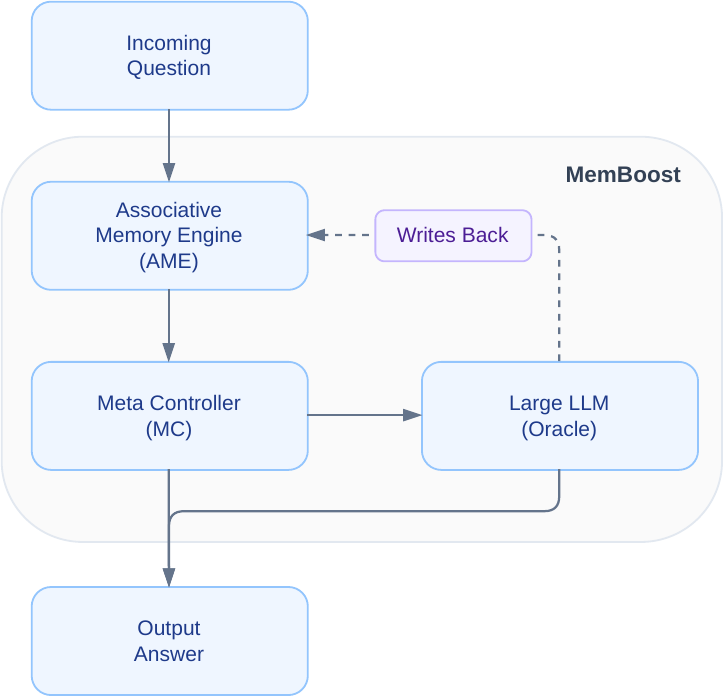}
    \caption{\textbf{Overview of MemBoost.} For each incoming query, the AME retrieves a set of relevant memory entries. The MC then either composes an answer from the retrieved results or escalates to the Oracle. When the Oracle is used, the MC decides whether to write the new answer back into the AME for future reuse.}
    \vspace{-6ex}
    \label{fig:memboost}
\end{figure}

As discussed above, in production settings, repeatedly invoking a frontier model to answer semantically redundant queries wastes substantial compute. To mitigate this inefficiency while balancing quality and cost, we propose MemBoost, a memory-boosted LLM serving system composed of three components (Figure~\ref{fig:memboost}): Associative Memory Engine, Large-LLM Oracle and Meta Controller.

\textbf{Associative Memory Engine (AME).}
AME maintains an external memory that stores auxiliary knowledge as well as previously answered queries and their associated responses, and supports fast semantic retrieval. Concretely, given a new query $x_t$ at time $t$, AME retrieves a small set of $K$ candidate memory entries \(
\mathcal{M}_t=\{(x^{(i)},y^{(i)},m^{(i)})\}_{i=1}^{K},
\) where $m^{(i)}$ denotes metadata (e.g., query category). The AME stores the question-answer pair for retrieval together with the qid (a unique identifier provided by the dataset), the category and the timestamp. In addition, AME supports write-back by storing newly generated high-quality entries $(x_t, y_t, m_t)$ when instructed by the Meta Controller.

\textbf{Large-LLM Oracle.}
The oracle is a high-capability model that produces high-quality answers but incurs high inference cost. We use it as a fallback: the system queries the oracle when retrieval from AME is missing, ambiguous, or deemed unreliable by the Meta Controller. We denote the oracle’s response to query $x_t$ as \(
y_t^\star=\mathrm{Oracle}(x_t).
\)

\textbf{Meta-Controller (MC).}
MC is a lightweight LLM that orchestrates the interaction among the user, AME, and the oracle. At each step, MC decides whether to answer using information retrieved from AME or to escalate to the oracle. When escalation occurs, MC also determines whether the newly generated query answer pair should be stored in AME for future reuse. Concretely, given a query $x_t$, MC first requests retrieval from AME and obtains $\mathcal{M}_t$. Conditioned on $x_t$ and $\mathcal{M}_t$, MC either returns an answer directly, $y_t=\mathrm{MC}(x_t,\mathcal{M}_t)$, or calls the oracle to obtain $y_t^\star=\mathrm{Oracle}(x_t)$ and returns $y_t=y_t^\star$. In the latter case, MC may optionally write $(x_t, y_t^\star, m_t)$ back into AME, where $m_t$ denotes metadata inferred by MC. This “retrieve $\rightarrow$ decide $\rightarrow$ escalate (if needed) $\rightarrow$ write-back (if needed)” loop reduces expensive oracle calls under repeated queries while preserving high answer quality when retrieval is insufficient.

\subsection{System Cost}
\label{sec:cost}

To model system efficiency,  let $c_O(x_t)$ denote the cost of an oracle call at time $t$ (e.g., GPU time, energy, or monetary cost), $c_M(x_t)$ the cost of running the lightweight MC, and $c_R(x_t)$ denote the cost of retrieval (e.g., CPU time). In typical deployments, $c_O(x_t)$ is much larger than $c_M(x_t)+c_R(x_t)$, since frontier-model
inference dominates GPU compute, whereas retrieval is primarily CPU-bound and
inexpensive relative to large-model generation. Let $I_t\in\{0,1\}$ indicate whether the retrieval information from AME is used at time $t$ ($I_t=1$ if memory is used, and $I_t=1$ if the system escalates to the oracle). The
total cost over $T$ steps is
\[
C_T
=\sum_{t=1}^{T}\bigl(c_M(x_t)+c_R(x_t)\bigr)
+\sum_{t=1}^{T} (1-I_t)\,c_O(x_t).
\]
By contrast, an oracle-only baseline would cost approximately $C_T^{\text{oracle}}=\sum_{t=1}^{T}c_O(x_t)$. So our framework achieves
cost savings whenever $C_T < C_T^{\text{oracle}}$. Equivalently, this condition can be written as
\[
\sum_{t=1}^{T} I_t\,c_O(x_t)
>
\sum_{t=1}^{T}\bigl(c_M(x_t)+c_R(x_t)\bigr).
\]
Putting quality and cost together, our goal is to achieve oracle-level quality with significantly
fewer oracle invocations:
\[
\frac{1}{T}\sum_{t=1}^{T}r(x_t,y_t)
\approx
\frac{1}{T}\sum_{t=1}^{T}r\bigl(x_t,y_t^\star\bigr), C_T < C_T^{\text{oracle}}
\]

\section{Experiments}
In this section, we evaluate MemBoost in terms of both (i) reducing inference cost and (ii) retaining high response quality.

\subsection{Experiment Setup}
We evaluate the proposed MemBoost framework on the MMLU-Pro dataset~\cite{wang2024mmlupro}, a widely used and challenging benchmark that covers diverse disciplines and is designed to more rigorously assess LLM capabilities. We use the Business category of the MMLU-Pro dataset and ground truth label answers as our benchmark for comparing the accuracy of the meta-controller (MC) with the Large LLM (Oracle). As the Business category contains 768 examples it is large enough to fit 5000 requests of the chosen Zipf distributions. To emulate real-world LLM-serving workloads where many requests are repeated or near-duplicated, we generate a query stream by sampling MMLU-Pro questions according to a Zipf distribution~\cite{zipf1949human}. This produces a heavy-tailed access pattern in which a small number of questions occur frequently while most questions appear rarely, capturing the repetition behavior commonly observed in practice. We vary the Zipf parameter ($\alpha$) to obtain different repetition rates and study how workload skew affects MemBoost. For the lightweight Meta Controller (MC), we evaluate several small-scale LLMs, including Qwen-3.5-2B~\cite{qwen2026qwen3_5_2b}, Ministral-3-3B-Instruct-2512~\cite{liu2026ministral3}, and Qwen3-4B-Instruct-2507-FP8~\cite{yang2025qwen3}. For the Large-LLM Oracle, we use Qwen3-14B-FP8-dynamic~\cite{yang2025qwen3,micikevicius2022fp8}. To support fast retrieval in the Associative Memory Engine (AME), we embed stored query-answer pairs using all-MiniLM-L6-v2~\cite{wang2020minilm,reimers2019sentencebert} and perform approximate nearest-neighbor search with a FAISS cosine-similarity index~\cite{douze2024faiss}. 

\begin{table}[t]
\centering
\scriptsize
\setlength{\tabcolsep}{4pt}
\renewcommand{\arraystretch}{0.95}
\resizebox{\columnwidth}{!}{%
\begin{tabular}{@{}llc@{}}
\toprule
\textbf{Component} & \textbf{Hyperparameter} & \textbf{Value} \\
\midrule
\textbf{Traffic simulation} 
& Workload distribution & Zipfian \\
& Zipf exponent ($\alpha$) & $\{0.8, 1.1, 1.4\}$ \\
& Number of requests ($N$) & 5{,}000 \\
& Random seed & 1 \\
\midrule
\textbf{Associative Memory Engine}
& Embedding model & \texttt{all-MiniLM-L6-v2} \\
& Similarity metric & Cosine (FAISS inner product) \\
& Similarity threshold ($\tau$) & 0.95 \\
& Retrieval top-$k$ & 3 \\
\midrule
\textbf{LLM generation (MC)}
& Temperature & 0.0 \\
& Max generation tokens & 4096 \\
& Frequency / presence penalty & 0.0 \\
& Chat template kwargs & \texttt{"enable\_thinking"=false} \\
\midrule
\textbf{LLM generation (Oracle)}
& Temperature & 0.0 \\
& Max generation tokens & 4096 \\
& Frequency / presence penalty & 0.0 \\
\midrule
\textbf{Environment}
& Python version & 3.10 \\
& NVIDIA driver & 573.57 \\
& CUDA version & 12.8 \\
\bottomrule
\end{tabular}}
\caption{System and hyperparameter configuration for the continuous serving simulation.}
\label{tab:hyperparameters}
\vspace{-6ex}
\end{table}

To ensure reproducibility of our experiment runs, we attach the exact hyperparameters as found in Table \ref{tab:hyperparameters} which we used across all experiments. All models were served using the vLLM library to optimize for throughput and latency \cite{kwon2023efficient}.
To ensure fair inference time comparisons, the Meta-Controller including the small LLM and the Associative Memory Engine and the Large LLM Oracle (solver) were each deployed on a dedicated NVIDIA A100 80GB GPU. Both the Meta-Controller and the Large LLM Oracle are configured with Temperature = $0.0$ to maintain deterministic responses. In addition the Oracle and the Router model had a 4,096 max tokens setting to allow for 5-shot Chain-of-Thought reasoning. We use the Business category of the MMLU-Pro dataset and ground truth label answers as our benchmark for comparing the accuracy of the meta-controller (MC) with the Large LLM (Oracle). As the Business category contains 768 examples it is large enough to fit 5000 requests of the chosen Zipf distributions.

\subsection{Preliminary Results}

\begin{table}[ht]
\centering
\scriptsize
\setlength{\tabcolsep}{3pt}
\renewcommand{\arraystretch}{0.95}
\begin{tabularx}{\columnwidth}{@{}>{\raggedright\arraybackslash}X
>{\centering\arraybackslash}p{0.11\columnwidth}
>{\centering\arraybackslash}p{0.11\columnwidth}
>{\centering\arraybackslash}p{0.11\columnwidth}@{}}
\toprule
\multirow{2}{*}{Model} & \multicolumn{3}{c}{Accuracy (\%)} \\
\cmidrule(lr){2-4}
& Zipf 0.8 & Zipf 1.1 & Zipf 1.4  \\
\midrule
\textit{Baselines} \\
Qwen3.5-2B & 50.0 & 43.5 & 37.1 \\
Ministral-3-3B-Instruct-2512 & 53.8 & 46.4 & 38.2 \\
Qwen3-4B-Instruct-2507-FP8 & 74.5 & 75.6 &  80.5 \\
Qwen3-14B-FP8-dynamic (Oracle) & 76.4 & 79.9 & 85.0 \\
\midrule
\textit{MemBoost (ours)} \\
\textbf{MemBoost (Qwen3.5-2B)} & \textbf{76.7} & \textbf{81.8} & \textbf{87.4} \\
MemBoost (Ministral-3-3B-Instruct-2512) & 76.2 & 79.7 & 85.0 \\
MemBoost (Qwen3-4B-Instruct-2507-FP8) & 76.1 & 79.8 & 85.0 \\
\bottomrule
\end{tabularx}
\caption{Accuracy (\%) of different methods on MMLU-Pro under Zipf-sampled query streams with varying repetition rates (Zipf $\alpha$).}
\label{tab:mmlu_pro_zipf_accuracy}
\vspace{-4ex}
\end{table}

Table~\ref{tab:mmlu_pro_zipf_accuracy} reports accuracy (\%) against the MMLU-Pro ground-truth labels for different methods under Zipf-sampled query streams with varying repetition rates (Zipf exponent $\alpha$). Across all settings, MemBoost consistently improves over the corresponding small-model baselines and achieves performance comparable to the oracle. Notably, MemBoost with Qwen3.5-2B even outperforms the oracle in all three workloads. We attribute this to MemBoost’s ability to reuse previously generated high-quality answers through semantic memory: once a question (or a near-duplicate) has been answered correctly and written back, subsequent occurrences can be served directly from memory, avoiding additional generation errors. Finally, MemBoost exhibits a clear improvement as the workload becomes more skewed (larger $\alpha$), since higher repetition rates lead to more memory hits, thereby increasing the fraction of queries answered using stored correct responses.


\begin{figure}[H]
    \centering
    \includegraphics[width=0.95 \linewidth]{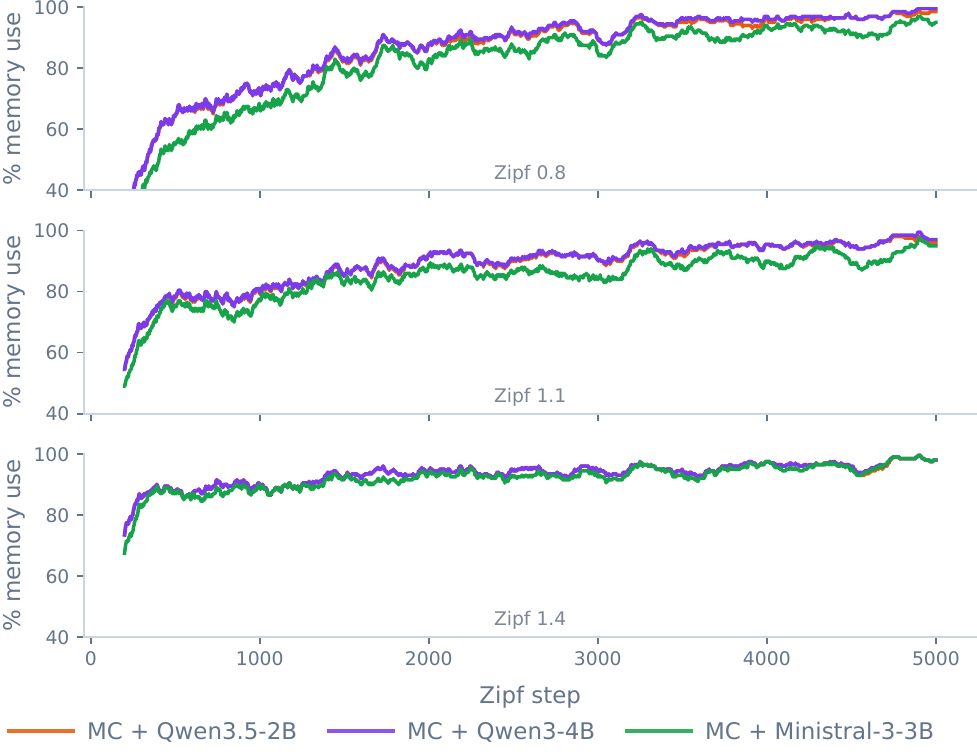}
    \caption{Average memory-use rate \(\overline{I}_t\) (200-step window) over a 5{,}000-step Zipf-sampled query stream. Higher \(\overline{I}_t\) indicates more queries served from AME and fewer oracle calls, implying lower total inference cost.}
    \label{fig:mem_over_time}
    \vspace{-2ex}
\end{figure}

To quantify the serving cost of MemBoost, we report the average memory-use rate over a 200-step window, i.e., \(
\overline{I}_t \;=\; \frac{1}{200}\sum_{s=t-199}^{t} I_s,
\)
where \(I_s=1\) indicates that the system answers using information retrieved from AME (and \(I_s=0\) indicates escalation to the oracle). Under our cost model, where the oracle cost \(c_O\) is assumed to dominate the Meta Controller and retrieval overhead \(c_M+c_R\), a larger \(\overline{I}_t\) corresponds to fewer oracle calls and thus lower total cost. Figure~\ref{fig:mem_over_time} plots \(\overline{I}_t\) over the 5{,}000-step workload for different MC choices and Zipf distributions. In all settings, \(\overline{I}_t\) increases over time as AME is progressively populated via the write-back mechanism, indicating that an increasing fraction of queries are served from memory rather than escalated to the oracle. Under our cost model, this implies that MemBoost achieves lower total cost than an oracle-only baseline. Moreover, memory usage is consistently higher under more skewed workloads (larger Zipf \(\alpha\)), where repeated queries occur more frequently, further reducing the number of expensive oracle calls.

\begin{figure}[t]
    \centering
    \includegraphics[width=0.95 \linewidth]{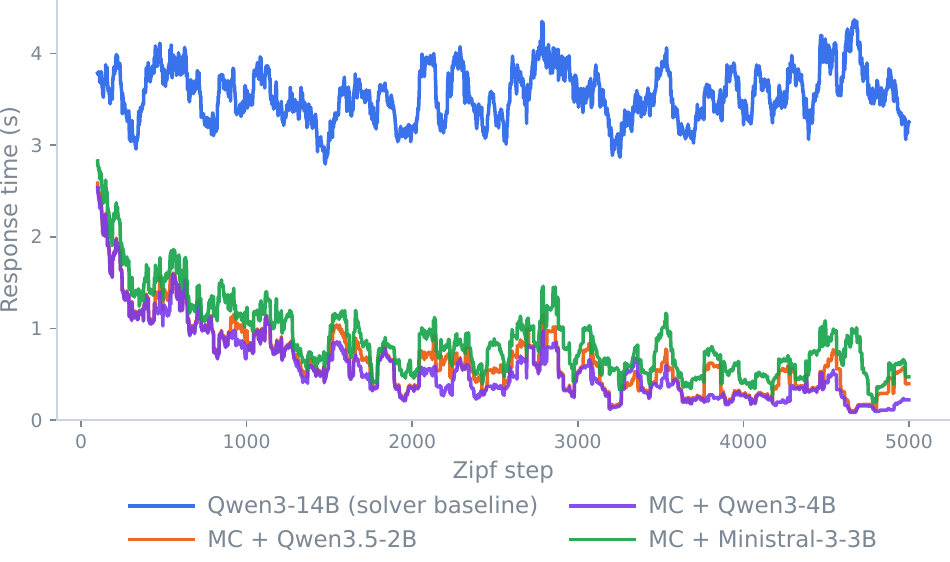}
    \caption{Response latency over time under Zipf-sampled workloads (average over the previous 100 steps). MemBoost reduces latency relative to the oracle-only baseline as an increasing fraction of queries are served from AME.}
    \label{fig:mem_vs_solver_latency}
    \vspace{-4ex}
\end{figure}

In addition to accuracy and system cost, we also report the latency of MemBoost. Figure~\ref{fig:mem_vs_solver_latency} shows the average response time over the previous 100 steps during the 5{,}000-step Zipf-sampled workload. As MemBoost increasingly answers queries using AME over time, the average latency steadily decreases and remains well below the oracle-only baseline. 

Overall, these results indicate that MemBoost is particularly effective for repeat-heavy workloads, preserving the strong answer quality of the larger oracle model while significantly reducing system cost and response latency. 


\section{Conclusion}
In this paper, we introduced \textbf{MemBoost}, a memory-boosted architecture for efficient LLM serving under interactive, repeat-heavy workloads. By combining an Associative Memory Engine, a high-capability Large-LLM Oracle, and a lightweight Meta-Controller for routing and response composition, MemBoost reduces redundant large-model inference while preserving answer quality. We hope MemBoost provides a practical foundation for building next-generation LLM services that jointly optimize quality, cost, and responsiveness.

\section*{Impact Statement}

This paper presents work whose goal is to advance the field of Machine
Learning. There are many potential societal consequences of our work, none
which we feel must be specifically highlighted here.

\bibliography{example_paper}
\bibliographystyle{icml2026}

\newpage
\appendix
\onecolumn

\section{Limitations}


While MemBoost shows promising accuracy and reduced inference cost on MMLU-Pro, the framework has not yet been evaluated in more diverse and demanding settings, such as open-ended long-form question answering or coding tasks, where outputs are longer, and error modes may differ. In addition, although Zipf sampling is used to simulate repeat-heavy workloads, the current simulation largely reflects repeated questions drawn from a fixed benchmark. In real deployments, queries are often not exact duplicates but only semantically similar, which can increase the risk of retrieval errors. A more realistic evaluation would therefore include workloads with paraphrased or semantically overlapping queries to stress-test robustness, particularly with respect to potential false hits in AME retrieval.

\section{Related Work}

\subsection{Retrieval-Augmented Language Modeling}

Retrieval-augmented generation (RAG) is a common way to improve language models by letting them look up relevant information from an external collection of documents, and then using that retrieved text to help produce an answer~\cite{lewis2020retrieval,liu2024much,du2024vul,mansurova2024qa}. Early work explored different ways to combine retrieval with generation, including training the retriever together with the language model~\cite{guu2020retrieval}, using retrieval to support open-domain question answering and other knowledge-intensive tasks \cite{lewis2020retrieval,izacard2023atlas}, and scaling retrieval to very large datastores to improve language modeling and downstream performance \cite{borgeaud2022improving}. A key advantage of these approaches is that knowledge can be updated by changing the document index, without having to retrain the model, and retrieval can improve factual coverage without simply making the model larger \cite{guu2020retrieval,lewis2020retrieval,izacard2023atlas}. More recent work studies how to make retrieval easier to use in practice. For example, some methods encourage the model to retrieve only when needed and to check whether the retrieved evidence actually supports the response \cite{asai2023self}. Overall, however, the main goal of RAG remains the same: improving the quality of a single response by grounding it in external evidence. As a result, standard RAG typically does not focus on reusing previously generated answers across users or sessions, nor does it explicitly support escalating to a stronger model when retrieval is insufficient.

\subsection{Semantic Caching}

Semantic caching extends classical caching to LLM services by reusing previous answers not only for exact duplicate queries, but also for queries that are semantically similar. In practice, a system stores past query-response pairs, represents queries with embeddings, and returns a cached response when a new query is close enough to an existing one, avoiding a full model call~\cite{bang-2023-gptcache,regmi2024gpt,li2024scalm,wang2025category}. Recent work has shown that this can significantly reduce latency and cost in real deployments, since many user requests concentrate on a small set of recurring intents \cite{bang-2023-gptcache,li2024scalm}. At the same time, semantic caching introduces new reliability challenges. Because a false hit can directly harm correctness (e.g., returning an answer for a different but similar-looking question), while a false miss loses potential savings~\cite{gill2025meancache}. This has motivated work on improving similarity matching and cache policies, including user-centric designs~\cite{gill2025meancache}, context-aware caching for multi-turn interactions~\cite{yu2025smartcache}, and more principled formulations of cache management and eviction under unknown workloads~\cite{liu2025semantic}. Overall, prior semantic caching research typically focuses on when to reuse cached outputs and how to manage the cache, but it often treats the fallback generation model and the caching layer as loosely coupled components~\cite{bang-2023-gptcache,li2024scalm}.

\subsection{Routing in LLMs}
A complementary line of work studies routing across multiple LLMs to reduce inference cost~\cite{chen2023frugalgpt,ong2024routellm,zhang2025router,wang2025mixllm,fedus2022switch}. The core idea is to maintain a pool of models with different cost-quality trade-offs and decide, for each query, which model to use. Many systems follow a cascade pattern: attempt a query with a cheaper model first and escalate to a stronger model only when needed, guided by a confidence or quality estimate~\cite{chen2023frugalgpt}. More recent approaches learn routing policies directly from preference data so that routing decisions better match human judgments of quality while reducing expensive model calls~\cite{ong2024routellm}. Beyond single-step routing, recent work also explores more adaptive and sequential routing strategies. For example, some methods treat routing as a multi-round decision process and train routers that interleave reasoning with routing actions~\cite{zhang2025router}. Other work considers routing in heterogeneous model pools where different models have complementary strengths, requiring robust routing strategies under distribution shift~\cite{wang2025mixllm}. While routing methods are effective at reducing calls to the strongest model, they typically assume that each query is still handled by some model generation pass and do not explicitly leverage cross-request answer reuse as a primary mechanism for efficiency~\cite{chen2023frugalgpt,ong2024routellm}. Separately, there is also extensive work on routing within a single model, such as mixture-of-experts architectures, which improves efficiency by activating sparse components but addresses a different level of the system stack~\cite{fedus2022switch}.


\end{document}